\documentclass[letterpaper, 10 pt, conference]{ieeeconf}  

\IEEEoverridecommandlockouts                              

\usepackage{subcaption}

\usepackage{amsmath}
\usepackage{amssymb}

\usepackage{tabularx}
\usepackage{graphicx}
\usepackage[font=footnotesize]{caption}
\usepackage{color}
\usepackage{hyperref}

\newcommand{\vecb}[1]{\boldsymbol{#1}}

\title{\LARGE \bf
Hierarchical Optimization-based Control for Whole-body Loco-manipulation of Heavy Objects
}

\author{Alberto Rigo, Muqun Hu, Satyandra K. Gupta, and Quan Nguyen
\thanks{This work is supported in part by National Science Foundation Grant IIS-2133091. The opinions expressed are those of the authors and do not necessarily reflect the opinions of the sponsors.}
\thanks{$^{1}$Alberto Rigo, Muqun Hu, Satyandra K. Gupta, and Quan Nguyen, are with the Department of Aerospace and Mechanical Engineering,
        University of Southern California, Los Angeles, CA, 90089
        {\tt\small rigo@usc.edu}, {\tt\small muqunhu@alumni.usc.edu}, {\tt\small quann@usc.edu}, {\tt\small guptask@usc.edu}}%
}

\begin{document}

\maketitle
\thispagestyle{empty}
\pagestyle{empty}

\begin{abstract}

In recent years, the field of legged robotics has seen growing interest in enhancing the capabilities of these robots through the integration of articulated robotic arms. However, achieving successful loco-manipulation, especially involving interaction with heavy objects, is far from straightforward, as object manipulation can introduce substantial disturbances that impact the robot's locomotion. This paper presents a novel framework for legged loco-manipulation that considers whole-body coordination through a hierarchical optimization-based control framework. First, an online manipulation planner computes the manipulation forces and manipulated object task-based reference trajectory. Then, pose optimization aligns the robot's trajectory with kinematic constraints. The resultant robot reference trajectory is executed via a linear MPC controller incorporating the desired manipulation forces into its prediction model. Our approach has been validated in simulation and hardware experiments, highlighting the necessity of whole-body optimization compared to the baseline locomotion MPC when interacting with heavy objects. Experimental results with Unitree Aliengo, equipped with a custom-made robotic arm, showcase its ability to lift and carry an 8kg payload and manipulate doors.

\end{abstract}

\section{INTRODUCTION}
Legged robots have garnered increasing attention for their potential to perform versatile locomotion tasks across challenging terrains \cite{di2018dynamic, jenelten2020perceptive, miki2022learning, kim2019highly, sombolestan2021adaptive}. To enhance their practical applicability, researchers have explored two approaches to perform loco-manipulation: utilizing the robot's body or existing limbs \cite{polverini2020multi, Rigo2023, sombolestan2023hierarchical} and integrating articulated robotic arms into legged platforms, enabling them to execute loco-manipulation tasks \cite{ferrolho2020optimizing, ma2022combining, morlando2022nonprehensile, xin2022loco}.  The combination of locomotion and manipulation capabilities opens up exciting opportunities for legged robots in various real-world applications, from search and rescue missions to industrial automation \cite{ferrolho2022roloma, rehman2016towards, gams2015accelerating, heppner2015laurope}.

However, achieving successful loco-manipulation in the context of legged robotics is a formidable challenge. Unlike traditional wheeled or tracked robots \cite{hou2018fast, woodruff2017planning, toussaint2018differentiable}, legged robots face unique dynamics and control intricacies when engaging in object manipulation. The introduction of manipulation forces and object interactions can introduce substantial disturbances that disrupt robot stability and locomotion \cite{li2023multi, mason1986mechanics}.

While recent years have witnessed increased attention towards learning-based methods for addressing loco-manipulation problems \cite{shi2021circus, sun2023bridging, ji2022hierarchical}, classic control frameworks remain prevalent. Loco-manipulation challenges entail coordinating manipulation actions with the robot's environment and locomotion. This typically involves two core components: a manipulation planner, often computing plans offline, and a locomotion tracking controller responsible for generating torque commands to track the high-level references established by the planner \cite{cebe2021online, dai2014whole, yao2022transferable, murphy2012high}. The tracking problem is typically tackled through various optimization frameworks that accommodate crucial system constraints \cite{zimmermann2021go, wolfslag2020optimisation}. In contrast, the planning module strives to strike a balance between physical accuracy and computational complexity, frequently relying on simplified models of the problem under consideration \cite{mittal2022articulated, bellicoso2019alma}. Nevertheless, exceptions exist where the full system dynamics are employed \cite{ewen2021generating, chiu2022collision}. However, a common limitation prevails: most of these formulations do not allow real-time solutions, rendering them unsuitable for rapid online replanning.

\begin{figure}
  \centering
  \includegraphics[width=0.92\linewidth,trim=100pt 130pt 130pt 50pt, clip]{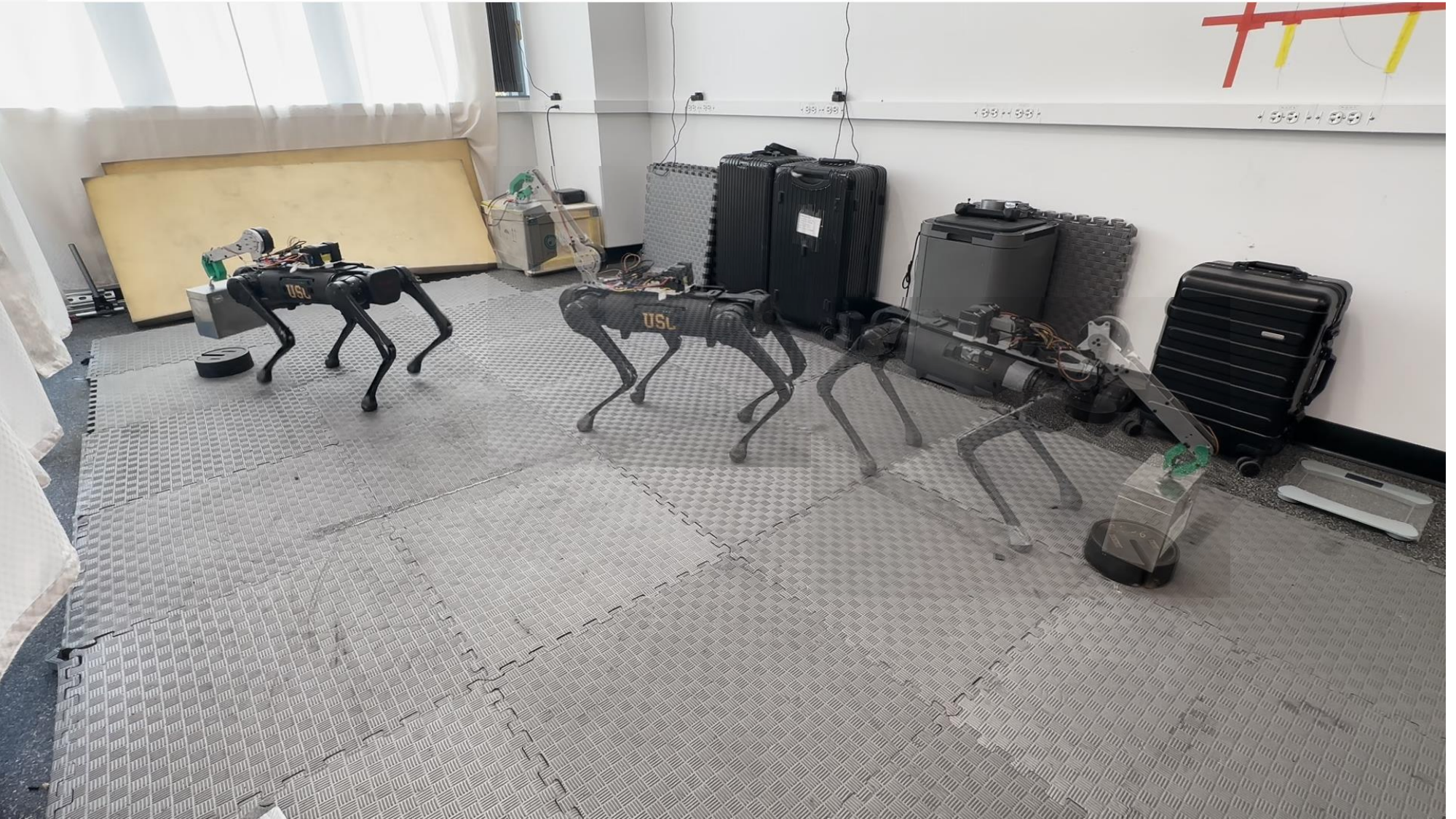}
  \caption{Snapshots of Aliengo lifting and carrying a 5kg payload. Supplemental video: \url{https://youtu.be/0hYDa94F78E}}
  \label{fig:intro}
\end{figure}

\begin{figure*}
  \medskip
  \centering
  \includegraphics[width=0.95\linewidth,trim=3pt 200pt 5pt 3pt, clip]{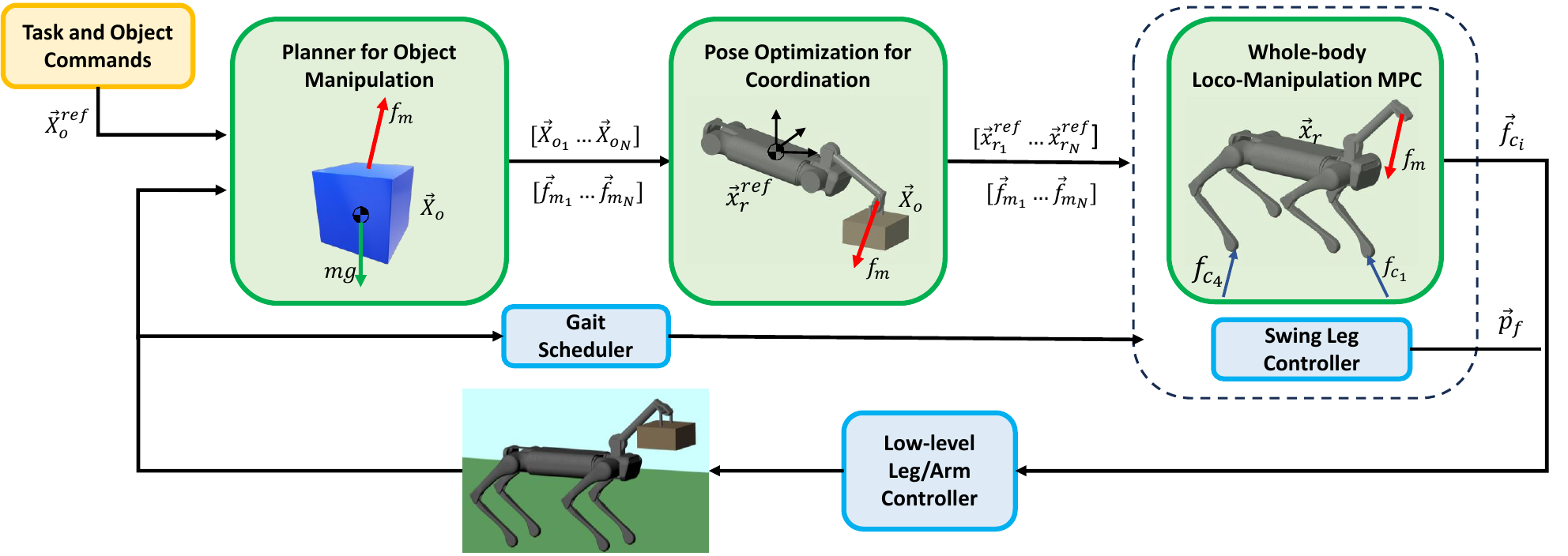}
  \caption{\textbf{Block diagram for the proposed framework}. Highlighted in green are the novel components that, together with the swing leg controller, form the high-level controller for the quadruped}
  \label{fig:archi}
\end{figure*}

Recent studies have adopted a unified non-linear loco-manipulation framework, such as \cite{sleiman2021unified} and \cite{sleiman2023versatile}. These works utilize a nonlinear Model Predictive Control planner, capable of computing real-time trajectories for the Center of Mass (CoM), limbs, and forces. A whole-body controller then tracks these reference trajectories. In \cite{sleiman2023versatile}, the authors introduce an offline planner that computes a sequence of locomotion and manipulation actions. Guided by a predefined library of interactions, this process facilitates the completion of user-defined tasks incorporating the environment model. However, it is important to note that while nonlinear MPC offers the advantage of considering more detailed predictive models or constraints, it typically requires significant computational power due to the complexity of the nonlinear optimization problem.

This paper presents a novel framework for legged agile loco-manipulation that leverages whole-body coordination to tackle the inherent complexities of the task. Our approach integrates elements from Model Predictive Control (MPC) and pose optimization to synthesize a control strategy that coordinates locomotion and manipulation, as illustrated in Fig. \ref{fig:intro}. We have devised a control structure capable of executing loco-manipulation tasks by decomposing the nonlinear problem into elementary components that interact hierarchically. Utilizing online pose optimization enables full coordination between manipulation and locomotion, demonstrated by the complex tasks we can perform with the custom-made 1-DOF robot arm introduced in this paper.

The remainder of the paper is organized as follows. Section \ref{sec:2} introduces the proposed control architecture explaining briefly the hierarchical interactions between the components. Each of these components is explained in detail in Sections \ref{sec:planner},\ref{sec:po},\ref{sec:mpc}. Then, Section \ref{sec:results} shows the results of simulation and hardware experiments.

\section{CONTROL SYSTEM OVERVIEW}
\label{sec:2}
In this section, we present the control system architecture, illustrated in Fig. \ref{fig:archi}, that underlies our proposed whole-body coordination framework. Loco-manipulation problems are usually described with nonlinear models due to the mutual interactions between robot and object. Nevertheless, solving nonlinear optimization problems is a computational burden and requires powerful on-board capabilities for the robot. We split the loco-manipulation problems into three elementary sub-problems that work hierarchically to circumvent this issue. Our approach begins by defining a user-specified task and the corresponding commands for the manipulated object. The initial phase of our method treats the object as an isolated entity subjected to manipulation forces. Using an MPC structure with a linear prediction model for object dynamics, we calculate the optimal object states and manipulation forces, aligning them with the desired manipulation task commands.
Subsequently, we introduce pose optimization, a critical step to coordinate the loco-manipulation. Pose optimization takes as input the desired object states, which we aim to track, and the manipulation force derived from the manipulation planner. It then generates a sequence of poses that define the robot's reference trajectory. Pose optimization also offers the advantage of enabling flexible loco-manipulation, accommodating various object parameter configurations, such as dimensions and gripping points.
With the robot's reference trajectory now defined through pose optimization, the whole-body loco-manipulation MPC efficiently tracks this trajectory while considering the impact of manipulation forces on the robot's dynamic stability.
The sections \ref{sec:planner}, \ref{sec:po}, \ref{sec:mpc} provide a detailed breakdown of each component of our proposed approach.

\section{PLANNER FOR OBJECT MANIPULATION}
\label{sec:planner}
The first component of our framework focuses on computing the necessary manipulation actions to achieve pre-defined tasks. We employ a linear Model Predictive Controller (MPC) structure that shares the same control horizon as the final robot controller, ensuring seamless coordination between manipulation and locomotion. Each manipulation task specifies the object states to be planned and the manipulation forces to be optimized. For instance, if the task involves opening a hinged door, we plan the opening angle and the force required to manipulate the handle. The chosen task also dictates the dynamics of the object under consideration. We can apply the same MPC planner to accommodate various tasks while adjusting the object states and manipulation forces accordingly.

The general dynamic formulation for the manipulated object in our linear MPC is expressed as follows:
\begin{equation}
     \vecb{A}_m\Dot{\vecb{X}}_{o} = \vecb{f}_{\mu} + \vecb{f}_m
    \label{eq:obj_dyn}
\end{equation}
Here, $\vecb{X}_{o}$ encompasses the object states we optimize, $\vecb{A}_m$ represents the diagonal matrix for the system linear dynamics, $\vecb{f}_{\mu}$ denotes external forces acting on the object (e.g., frictional forces), and $\vecb{f}_m$ stands for the manipulation force exerted on the object.

Different tasks involve different object dynamics and the corresponding commands. For example, lifting an object may require a command in terms of lifting velocity or lifting time.
The desired command for each task produces reference values for the states of the objects, denoted as $\vec{X}_o^{\text{ref}}$, which are incorporated into the cost function of the linear MPC problem to minimize deviations from these references.
We also include equality and inequality constraints tailored to the specific task at hand, such as guaranteeing that the manipulation force maintains its perpendicular orientation to the door surface while opening the door.

The desired manipulation force for the entire time horizon is passed to the quadruped loco-manipulation MPC and the pose optimization for whole-body coordination. In the former, the force is treated as a known quantity representing the interaction between the robot and the object over the prediction horizon. In the latter, the force is used to compute the arm joint torques needed to manipulate the object and minimize them. More details about how to use these forces are presented in the respective sections.
The optimized object trajectory from the MPC planner plays a crucial role in the subsequent pose optimization phase, where it coordinates the robot's locomotion to execute the desired manipulation task effectively.

\section{POSE OPTIMIZATION FOR COORDINATED LOCO-MANIPULATION}
\label{sec:po}
In this section, we present the details of pose optimization and its role in bridging the gap between the object manipulation planner and the whole-body loco-manipulation controller. The fundamental idea is to translate the computed manipulation forces and optimal object states into robot-centric states and dynamics. To achieve this, we perform optimized pose computations for the robot at each MPC horizon, accounting for the manipulation force and system kinematic constraints. Our approach is inspired by the work in \cite{li2023kinodynamics}, with the necessary modifications to adapt it to real-time execution while maintaining meaningful constraints.
The pose optimization problem is formulated as a Non-Linear Programming (NLP) problem, with the optimization variable $\vecb{X}$ encompassing the robot's Center of Mass (CoM) location $\vecb{p}_r$, body Euler angles $\vecb{\Theta}$, arm joint angles $\vecb{q}_{arm}$, and the manipulation force $\vecb{f_m}$ acting at the end-effector location. However, we exclude the leg joint angles and ground reaction forces from the optimization variables, which are computed by the swing leg controller and loco-manipulation MPC, allowing the solution of the NLP problem to be computed in real-time.
The problem is defined as follows:
\begin{align}
    \min_{\vecb{X}} \quad Q_p({p}_{r,z}&-{p}_{r,z}^{\textit{ref}})^2 + \lVert\vecb{\Theta}\rVert^2_{\vecb{Q}_\Theta} + \lVert\vecb{\tau}\rVert^2_{\vecb{Q}_\tau} \label{eq:nlp_cost} \\
    \text{s.t.} \quad & p_{z,min} \leq p_{hip_z}^i \leq p_{z,max} \label{eq:pz}\\
    & \vecb{\Theta}_{min} \leq \vecb{\Theta} < \vecb{\Theta}_{max} \label{eq:theta}\\
    & \vecb{q}_{arm, min} \leq \vecb{q}_{arm} \leq \vecb{q}_{arm, max} \label{eq:qarm}\\
    & \vecb{X}_e = \vecb{X}_{obj} \label{eq:Xe}\\
    & \vecb{f}_m = \vecb{f}_m^{\textit{plan}} \label{eq:fm}
\end{align}
Starting from eq. \ref{eq:nlp_cost}, the objectives of pose optimization are minimizing the difference between robot CoM height and reference value, minimizing the body rotations, and minimizing the arm torques $\vecb{\tau}$ needed for manipulation, calculated using the contact Jacobian of the arm $\vecb{J}_{arm}(\vecb{X})$.  These objectives are weighted by respective scalar or diagonal matrices $Q_p$, $\vecb{Q}_{\Theta}$, $\vecb{Q}_{\tau}$ of appropriate dimensions, allowing for tailored control of the robot's behavior during manipulation. While the CoM height is directly incorporated into the cost function, the $x-y$ position is determined by the optimization process based on problem constraints.  We avoid explicitly including leg joint angles in the pose, relying instead on Equation \ref{eq:pz} to ensure that each hip location's height remains within feasible bounds for effective stepping. Equation \ref{eq:theta} restricts robot orientation to physically feasible values, particularly concerning pitch and roll, while Equation \ref{eq:qarm} enforces constraints on the arm joint angles.
The last two constraints (Equations \ref{eq:Xe} and \ref{eq:fm}) establish the crucial connections between pose optimization and the MPC manipulation planner. Equation \ref{eq:Xe} ensures that the end effector's pose $\vecb{X}_e = \vecb{X}_e(\vecb{X})$ matches the optimized object states computed by the manipulation planner. The specifics of this constraint depend on the particular task at hand; for instance, in the case of lifting an object, it constrains the end effector's position, whereas for turning a door handle, it also includes the end effector's orientation. Equation \ref{eq:fm} requires that the manipulation force in the optimization variable aligns with the force computed by the manipulation planner $\vecb{f}_m^{\textit{plan}}$.
The pose optimization is executed at every MPC horizon, producing a reference trajectory for the robot's states for the whole-body loco-manipulation MPC.

\section{WHOLE-BODY LOCO-MANIPULATION MPC}
\label{sec:mpc}
In this section, we present the formulation of our proposed Loco-Manipulation Model Predictive Controller (MPC). We have developed this formulation to address a crucial aspect of our work: the manipulation of heavy objects, which significantly affects the dynamics of the robot. While the baseline locomotion MPC for quadruped robots, as introduced in \cite{di2018dynamic}, is designed to handle minor external disturbances to the robot's dynamics, it falls short when dealing with heavy objects. These disturbances need explicit consideration in the robot's dynamics.
Our model considers a single rigid body with contact forces applied at the feet locations and a manipulation force at the arm gripper, as illustrated in Fig. \ref{fig:archi}. We express the dynamic model in terms of the robot's position $\vecb{p_r}\in \mathbb{R}^3$ and angular velocity $\vecb{\omega_r}\in \mathbb{R}^3$, both in the world frame:
\begin{align}
    m_r(\vecb{\Ddot{p}}_r+\vecb{g}) &= \sum_{i=1}^{4}\vecb{f}_{c_i} + \vecb{f}_m ,  \label{eq:dyn_1}\\
    \vecb{I}_r\vecb{\Dot{\omega}}_r &= \sum_{i=1}^{4}\vecb{r}_{c_i} \times\vecb{f}_{c_i} + \vecb{r}_{m} \times\vecb{f}_{m} \label{eq:dyn_2} 
\end{align}
In Equation \ref{eq:dyn_1}, $m_r$ represents the combined mass of the robot and arm, $\vecb{g} = \begin{bmatrix}0 & 0 & \text{-}g\end{bmatrix}^T$ is the gravity vector, $\vecb{f}_{c_i}$ denotes the ground reaction force acting on foot $i$, and $\vecb{f}_m$ is the external manipulation force. In Equation \ref{eq:dyn_2}, the derivative of the angular momentum is simplified retaining only the term $\vecb{I}_r\vecb{\Dot{\omega}}_r$, and $\vecb{I}_r$ is the moment of inertia of the robot in the world frame, $\vecb{r}_{c_i}$ is the position vector from the robot's Center of Mass (CoM) to the $i^{\textit{th}}$ foot location, and $\vecb{r}_m$ is the position vector from the robot's CoM to the gripper location.

The manipulation force $\vecb{f}_m$ is a known quantity derived from the MPC manipulation planner. It represents the force the robot requires to follow the desired commands for various tasks. Furthermore, during the discretization of the dynamics for prediction in the MPC, we have access to the desired manipulation forces for all prediction horizons. Thus, we treat it as part of the input vector in the dynamics and enforce equality constraints at each horizon to match the force to the desired manipulation force.
With the prediction model defined, we represent it in a state-space form $\Dot{\vecb{x}}_r=\vecb{A}\vecb{x}+\vecb{B}\vecb{u}$, where:
\begin{align}
    \vecb{x_r} &= \begin{bmatrix} \vecb{\Theta} & \vecb{p}_r & \vecb{\omega}_r & \vecb{\Dot{p}}_r & g\end{bmatrix} \label{eq:x}\\
   \vecb{u} &= \begin{bmatrix} \vecb{f}_{c_1} & \vecb{f}_{c_2} & \vecb{f}_{c_3} & \vecb{f}_{c_4} & \vecb{f}_m\end{bmatrix} \label{eq:u}\\
    \vecb{A} &= \begin{bmatrix}
    \vecb{0}_{3} & \vecb{0}_{3} & \vecb{R}_z(\psi) & \vecb{0}_{3} & \vecb{0}_{3\times1}\\ \vecb{0}_{3} & \vecb{0}_{3} & \vecb{0}_{3} & \vecb{I}_{3} &  \vecb{0}_{3\times1}\\ \vecb{0}_{3} & \vecb{0}_{3} & \vecb{0}_{3} & \vecb{0}_{3} &  \vecb{0}_{3\times1} \\ \vecb{0}_{3} & \vecb{0}_{3} & \vecb{0}_{3} & \vecb{0}_{3} & \frac{\vecb{g}}{||\vecb{g}||}\\ \vecb{0}_{1\times3} & \vecb{0}_{1\times3}  & \vecb{0}_{1\times3}  & \vecb{0}_{1\times3} & 0\end{bmatrix} \label{eq:A}\\
    \vecb{B} &= \begin{bmatrix} 
    \vecb{0}_{3} & \cdots & \vecb{0}_{3} & \vecb{0}_{3}\\
    \vecb{0}_{3} & \cdots & \vecb{0}_{3} & \vecb{0}_{3} \vspace{5pt}\\
    \vecb{I}_r^{-1}\vecb{r}_{c_1}\times & \cdots & \vecb{I}_r^{-1}\vecb{r}_{c_4}\times & \vecb{I}_r^{-1}\vecb{r}_{m}\times \vspace{5pt}\\
    \frac{\vecb{I}_3}{m_r} & \cdots & \frac{\vecb{I}_3}{m_r} & \frac{\vecb{I}_3}{m_r}\\
    \vecb{0}_{1\times3} & \cdots & \vecb{0}_{1\times3} & \vecb{0}_{1\times3}
    \end{bmatrix} \label{eq:B}
\end{align}
In equation \ref{eq:A}, matrix $\vecb{R}_z(\psi)$ is the rotation matrix corresponding to the yaw angle $\psi$, while, in equation \ref{eq:B}, $\vecb{r}_i\times$ represents the skew-symmetric transformation matrix of position vector $\vecb{r}_i\in\mathbb{R}^3$. We can discretize the state-space formulation to use it as a prediction model in a linear MPC formulation, with $N$ horizons, defined as follows:
\begin{align}
    \min_{\vecb{x}_{r},\vecb{u}} \quad \sum_{i=1}^{N} &\lVert{\vecb{x}_{r}}_{i+1} - \vecb{x}_{{r}_{i+1}}^{\textit{ref}}\rVert^2_{\vecb{Q_r}} + \lVert\vecb{u}_i\rVert^2_{\vecb{R_r}} \label{eq:cost}\\
    \text{s.t.} \quad &\vecb{x}_{r_{i+1}} = \vecb{A}_d\vecb{x}_{r_i} + \vecb{B}_d\vecb{u}_i\label{eq:dyn_constr}\\
    & \underline{\vecb{c}}_f \leq \vecb{C}\vecb{u}_i \leq \overline{\vecb{c}}_f \label{eq:fric_constr}\\
    & \vecb{f}_{c_i} = 0 \quad \text{if swing leg} \label{eq:swing_constr}\\
    & \vecb{f}_m = \vecb{f}_{m_{\textit{des}}} \label{eq:fm_constr} 
\end{align}
In the cost function of the problem, equation \ref{eq:cost}, $\vecb{x}_{r}^{\textit{ref}}$ denotes the reference values for the robot states, which are computed from the pose optimization presented in section \ref{sec:po}. These reference values are crucial for coordinating the entire-body motion to execute the loco-manipulation task accurately. Equation \ref{eq:dyn_constr} represents the dynamic constraints for each prediction horizon, and $\vecb{A_d}$ and $\vecb{B_d}$ are the discrete-time equivalents of the matrices presented in equations \ref{eq:A} and \ref{eq:B}. Equation \ref{eq:fric_constr} represents the frictional pyramid constraints for each leg, equation \ref{eq:swing_constr} enforces the vanishing of the reaction forces on the swing legs, and equation \ref{eq:fm_constr} is the equality constraint that sets the manipulation force in the dynamics equal to the desired manipulation force from the MPC manipulation planner for each prediction horizon.
The controller determines optimal ground reaction forces while accounting for the presence of the manipulation force at each prediction horizon, enabling precise tracking of desired state trajectories generated by the pose optimization.

\section{RESULTS}
\label{sec:results}
\begin{figure}
  \medskip
  \centering
  \includegraphics[width=0.9\linewidth,trim=0pt 0pt 0pt 0pt,clip]{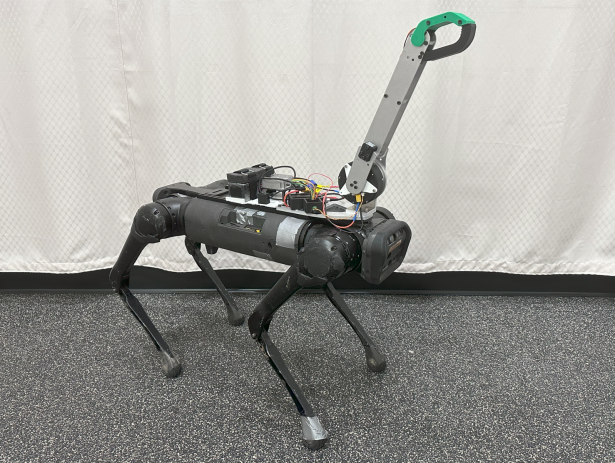}
  \caption{\textbf{Unitree Aliengo with custom-made arm} used for experimental validation of the proposed approach}
  \label{fig:real_robot}
\end{figure}

\begin{figure}
     \medskip
     \centering
     \begin{subfigure}[b]{0.47\linewidth}
         \centering
         \includegraphics[width=\textwidth,trim=0pt 0pt 0pt 10pt,clip]{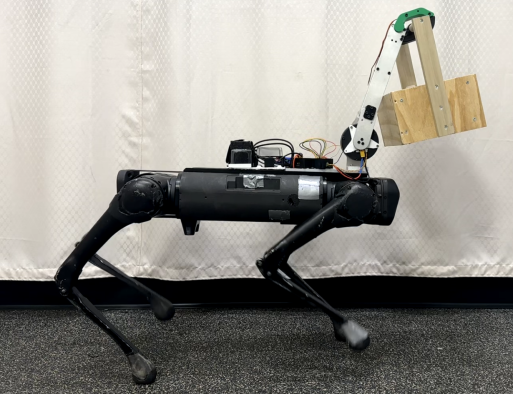}
         \caption{Snapshot of lifting 3Kg object using loco-manipulation MPC}
         \label{fig:lm_comp_5kg}
     \end{subfigure}
     \begin{subfigure}[b]{0.47\linewidth}
         \centering
         \includegraphics[width=\textwidth]{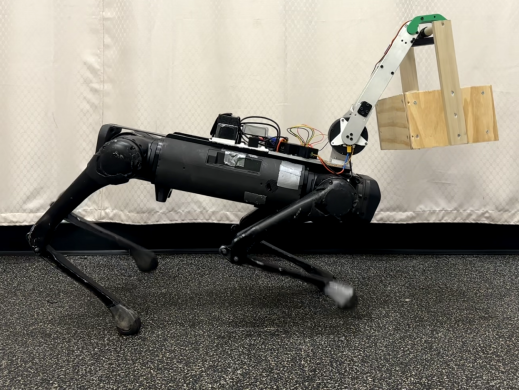}
         \caption{Snapshot of lifting 3Kg object using baseline MPC}
         \label{fig:base_comp_5kg}
     \end{subfigure}
     \begin{subfigure}[b]{0.96\linewidth}
        \vspace{2em}
         \centering
         \includegraphics[width=\textwidth]{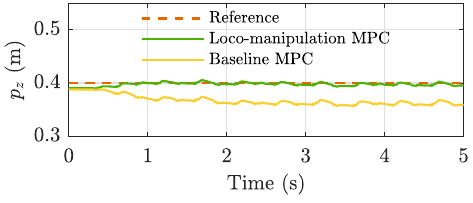}
         \caption{Robot COM height}
         \label{fig:comp_pz}
     \end{subfigure}
     \begin{subfigure}[b]{0.96\linewidth}
         \centering
         \includegraphics[width=\textwidth]{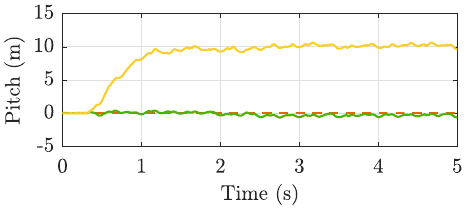}
         \caption{Robot Pitch}
         \label{fig:comp_pitch}
     \end{subfigure}
        \caption{\textbf{Improvements using loco-manipulation MPC}. In the plots, we compare the results using the loco-manipulation MPC and the baseline MPC. The robot is lifting from the ground a 3Kg object and reaches a predefined arm configuration,}
        \label{fig:5kg_comp}
\end{figure}

\begin{figure}
     \medskip
     \centering
     \begin{subfigure}[b]{0.96\linewidth}
         \centering
         \includegraphics[width=\textwidth]{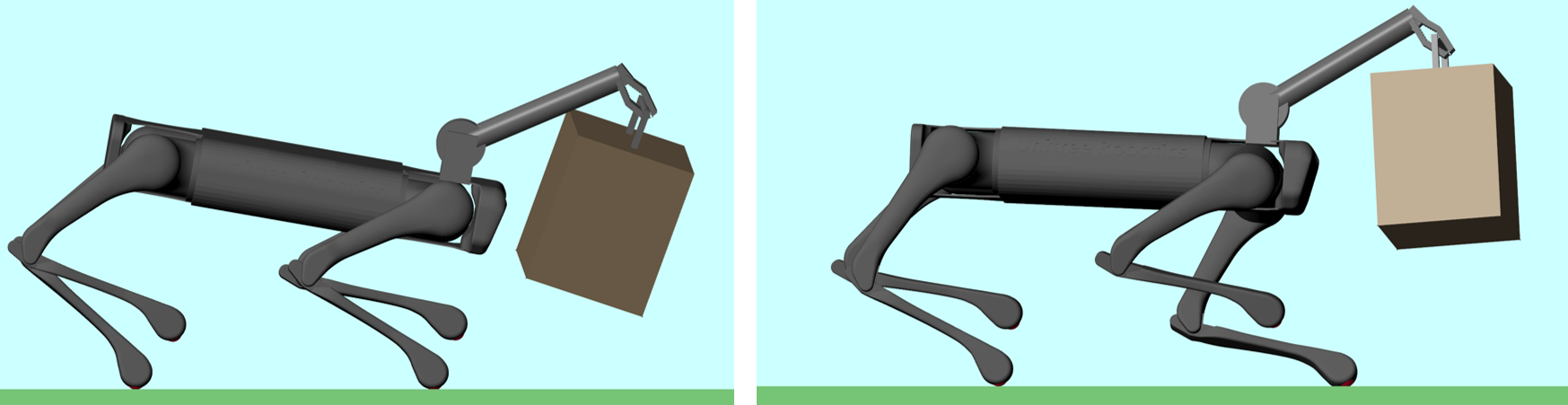}
         \caption{Using object manipulation planner}
         \label{fig:dl_planner_pz}
     \end{subfigure}
     \begin{subfigure}[b]{0.96\linewidth}
         \vspace{0.5em}
         \centering
         \includegraphics[width=\textwidth]{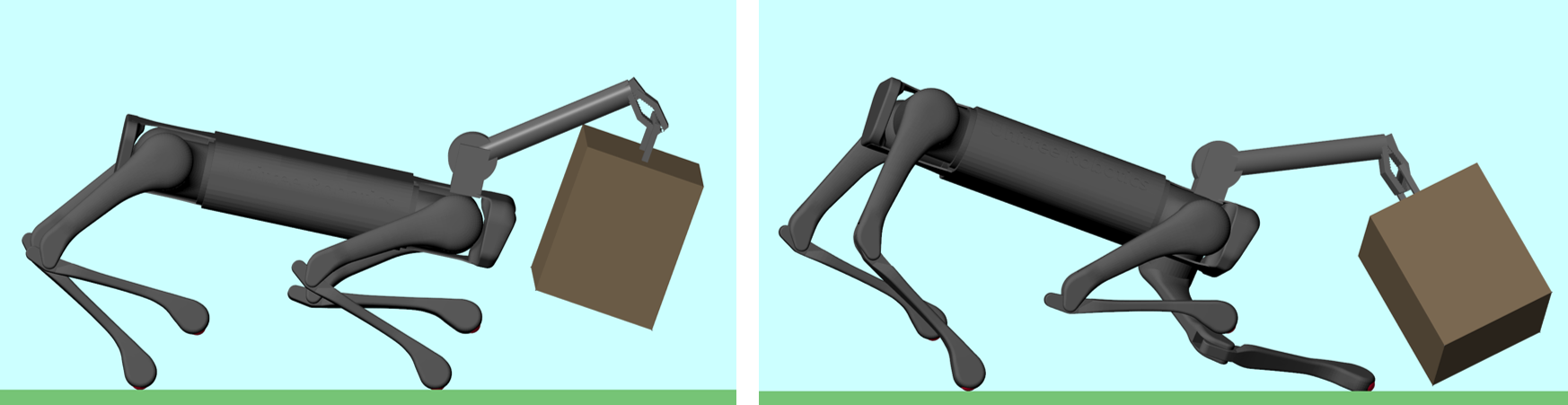}
         \caption{Using fixed manipulation force}
         \label{fig:dl_planner_pitch}
     \end{subfigure}
     \begin{subfigure}[b]{0.96\linewidth}
        \vspace{0.5em}
         \centering
         \includegraphics[width=\textwidth]{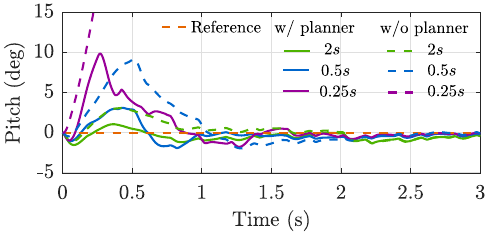}
         \caption{Pitch angle of the robot during the dynamic lifting task}
         \label{fig:dl_base_pz}
     \end{subfigure}
     \begin{subfigure}[b]{0.96\linewidth}
        \vspace{0.5em}
         \centering
         \includegraphics[width=\textwidth]{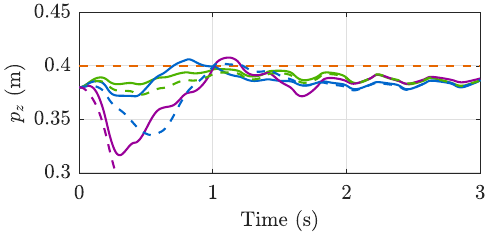}
         \caption{Height of the robot COM during the dynamic lifting task}
         \label{fig:dl_base_pitch}
     \end{subfigure}
        \caption{\textbf{Improvements using object manipulation planning}. In the plots, we compare the results using the loco-manipulation MPC with the object manipulation planner or a fixed manipulation force without the object manipulation planner. The robot lifts a 10 Kg object from the ground and reaches a predefined arm configuration in shorter times, making the lifting more dynamic.}
        \label{fig:dl_comp}
\end{figure}
In this section, we present the results we obtained in both simulation and real hardware to prove the effectiveness of our approach. Simulations were performed using the Simscape multibody package in Matlab Simulink, a high-fidelity environment that accurately simulates contact-rich scenarios. 
To test and validate our framework, we conducted experiments using a Unitree Aliengo robot equipped with a custom 1-DOF robotic arm, shown in Fig.\ref{fig:real_robot}. We designed this arm with three key considerations:
\begin{itemize}
    \item By limiting it to 1-DOF, we reduced the arm's weight, enabling a higher maximum payload capacity.
    \item By installing a single powerful actuator at the base of the arm, we can perform highly dynamic tasks with heavy objects.
    \item With proper whole-body coordination, we can still achieve many manipulation tasks even with the reduced arm DOFs.
\end{itemize}

Our control architecture, detailed in Section \ref{sec:2}, efficiently operates on the robot's onboard computer. The low-level controller operates at a frequency of 1 kHz. At the same time, the object manipulation planner and loco-manipulation Model Predictive Control (MPC) run at 30 Hz with a time horizon of $T = 0.5$ s and $N = 10$ horizons. We utilize CasADi and the Ipopt solver to solve the pose optimization problem. Despite the hierarchical structure of the framework leading to a higher number of hyperparameters, the linear formulations facilitate straightforward tuning. Through a series of experiments, we demonstrate the impact of each component of our approach compared to baseline controllers.

\subsection{Effect of Loco-manipulation MPC}

In this section, we want to highlight the importance of considering the object in the robot MPC model, especially when the object is heavy. To do this, we compare the loco-manipulation MPC in our approach to a baseline locomotion MPC, where there is no information about the object the robot is carrying. We perform this task on the real robot and during the task, the robot picks up a 3kg object from the ground using the arm DOF and lifts it to a predefined arm angle in 2 seconds. We can see snapshots of the task and COM height and robot pitch tracking with the two controllers in Fig. \ref{fig:5kg_comp}. Using the proposed loco-manipulation MPC, the robot can successfully lift the heavy object from the ground and maintain the desired height and pitch for the robot throughout the entire task. Instead, the baseline MPC cannot follow the desired quantities and struggles to maintain balance. In fact, with objects heavier than 5kg, the baseline MPC would fail, while our proposed controller can still handle them. Due to the mass-efficient arm design, we have a maximum payload capability of 8 kg, which is almost $50\% $ of the robot's weight. Our payload-to-robot weight ratio is higher than other robotic arms used in state-of-the-art legged loco-manipulation research. 

\begin{figure}
     \medskip
     \centering
     \begin{subfigure}[b]{0.48\linewidth}
         \centering
         \includegraphics[width=\textwidth,trim=0pt 40pt 0pt 120pt, clip]{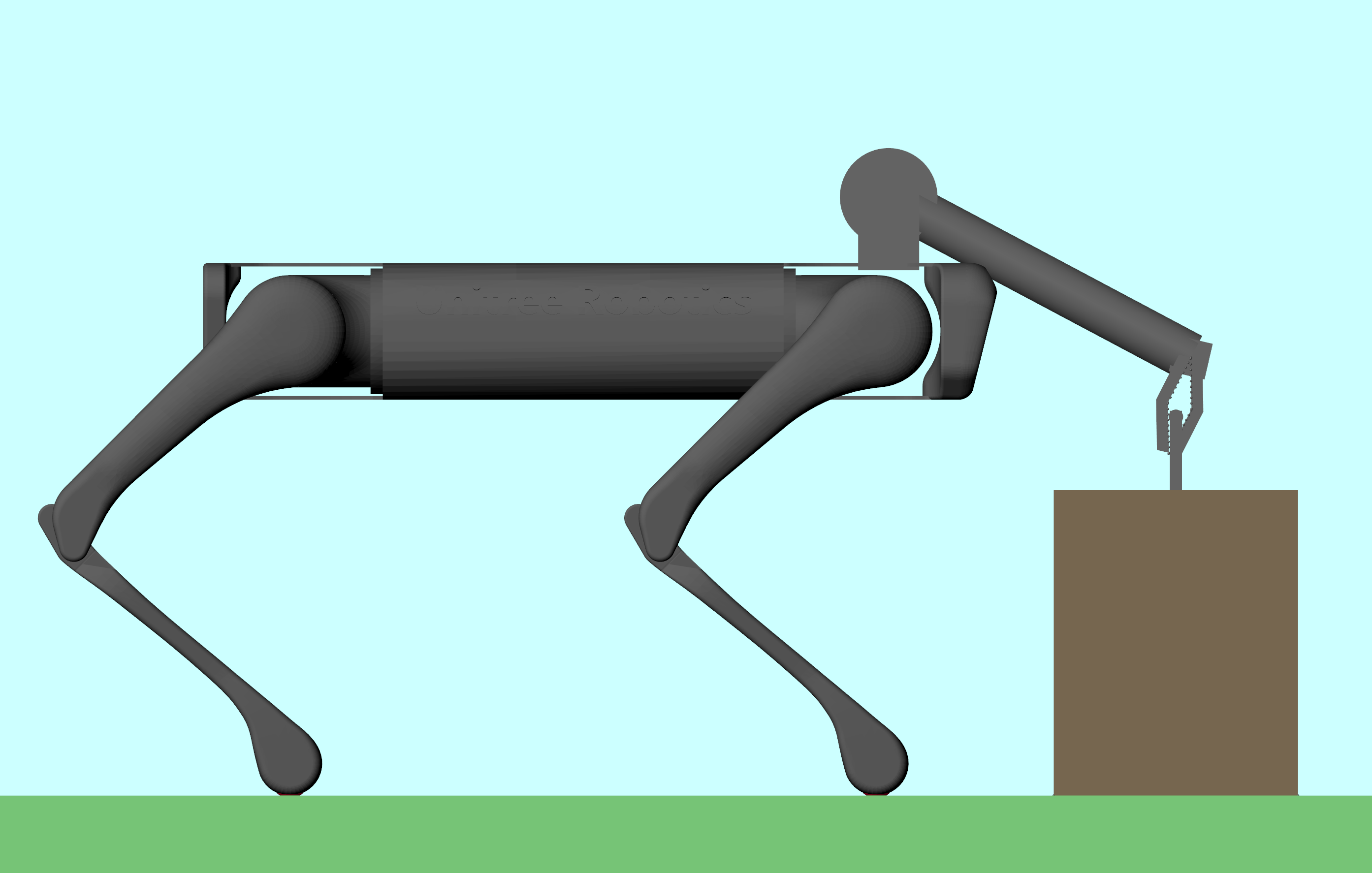}
         \caption{Baseline pose}
         \label{fig:po_base}
     \end{subfigure}
     \begin{subfigure}[b]{0.48\linewidth}
         \centering
         \includegraphics[width=\textwidth,trim=0pt 40pt 0pt 100pt, clip]{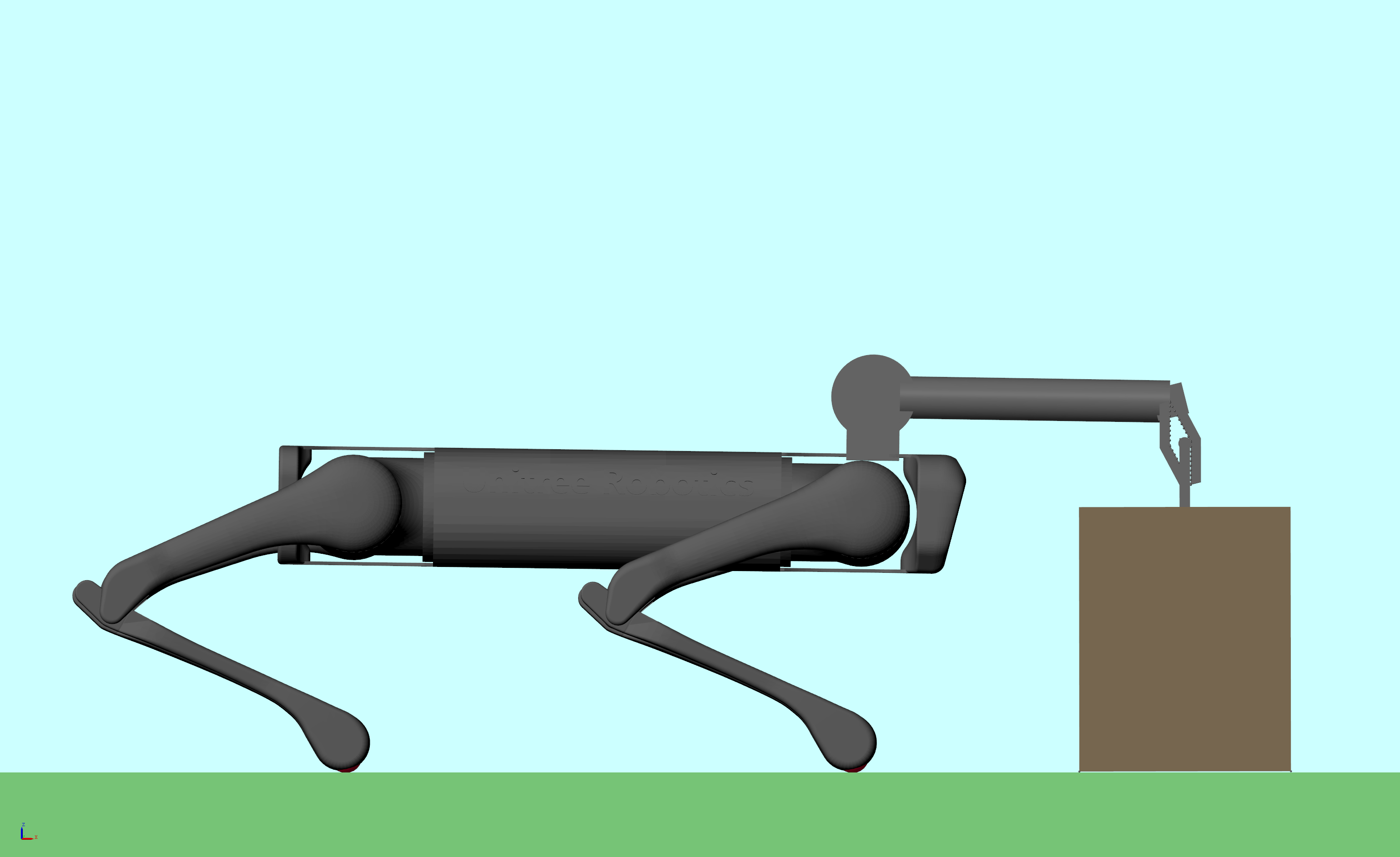}
         \caption{Squatted pose}
         \label{fig:po_squat}
     \end{subfigure}
     \begin{subfigure}[b]{0.48\linewidth}
        \vspace{0.5em}
         \centering
         \includegraphics[width=\textwidth,trim=0pt 40pt 0pt 120pt, clip]{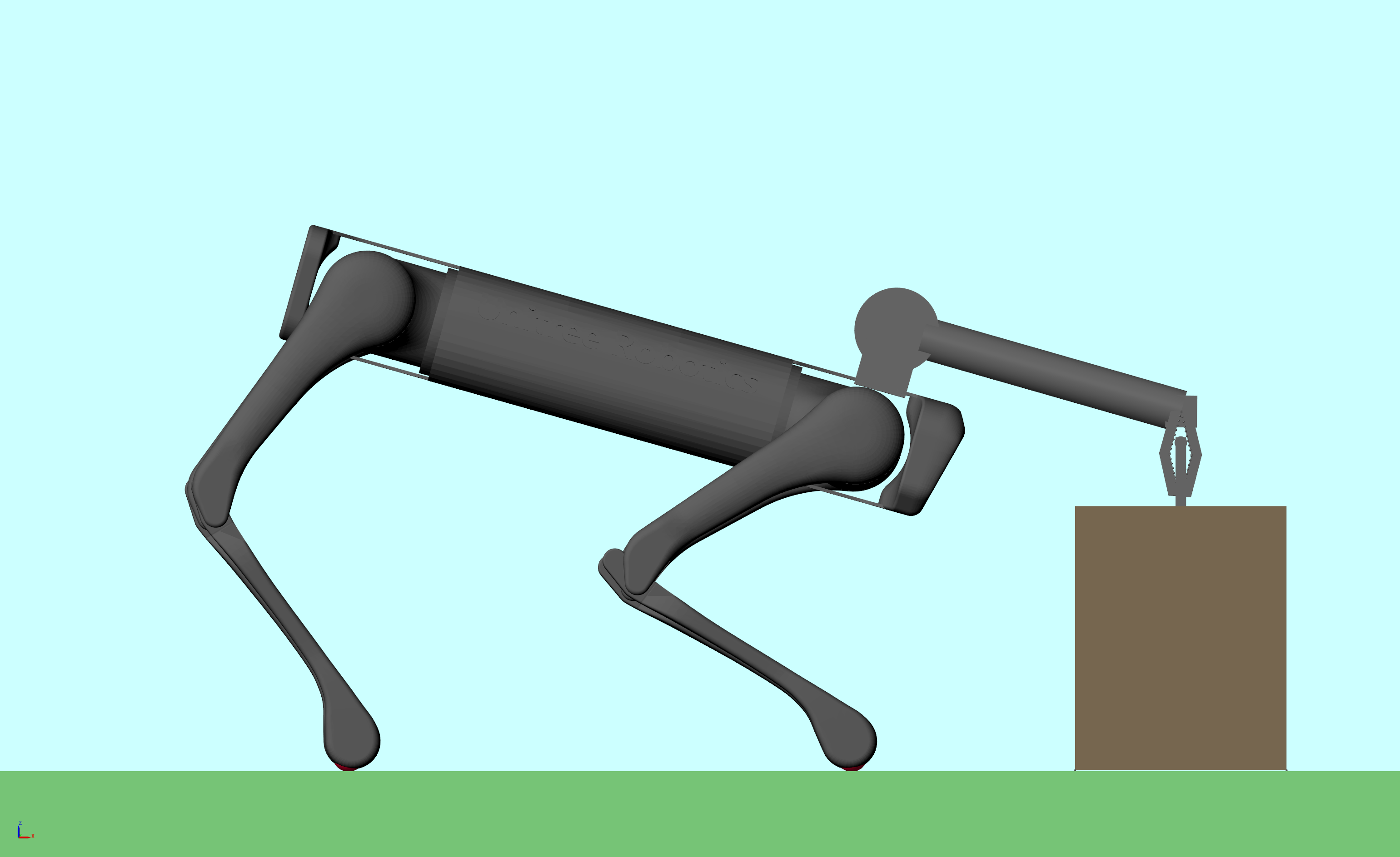}
         \caption{Pitched pose}
         \label{fig:po_pitch}
     \end{subfigure}
     \begin{subfigure}[b]{0.48\linewidth}
         \centering
         \includegraphics[width=\textwidth,trim=0pt 40pt 0pt 120pt, clip]{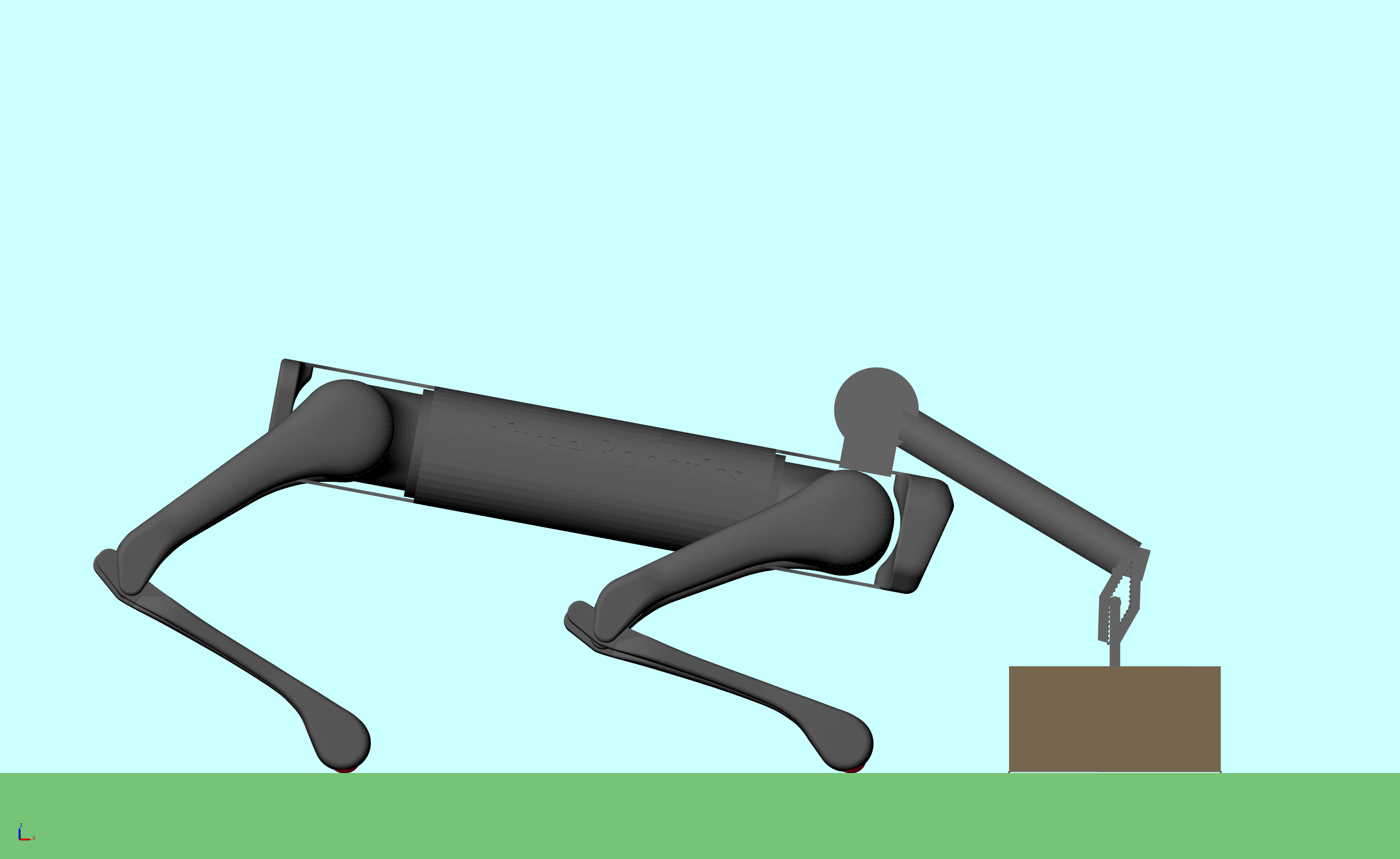}
         \caption{Mix pose for low object picking}
         \label{fig:po_mix}
     \end{subfigure}
        \caption{\textbf{Various optimized poses for object lifting}. These figures represent the optimal poses computed for the task based on the object dimensions and the weights in the optimization cost function.}
        \label{fig:po_comp}
\end{figure}
\subsection{Effect of Planner for Object Manipulation}

We then investigated the effectiveness of the planner for object manipulation. This component of the proposed approach becomes critical when performing very dynamic tasks because the effect of the object on the robot cannot be represented by a fixed manipulation force anymore. To highlight the importance of the component, we performed different simulations of dynamic object-lifting tasks, where various lifting times were specified for each case. The results are presented in Fig. \ref{fig:dl_comp} regarding robot COM height and pitch angle, using the object planner or without the planner with fixed manipulation force, together with the loco-manipulation MPC. We start from a 2-second lifting that can be treated as a quasi-static movement, and both simulations are successful. But if we reduce the lifting time to $0.5s$, we start to see that the controller that uses the fixed manipulation force cannot maintain the robot's reference height and pitch. When the lifting time decreases to $0.25s$, this controller fails, while our proposed approach using the manipulation planner can consider the increased force acting on the robot from the large acceleration imparted on the object and still maintain balance to complete the task.

\subsection{Effect of Pose Optimization for Coordination}

The next component we want to showcase is the pose optimization for coordinated loco-manipulation. This part of the approach is important because it links the output of the object manipulation planner to the loco-manipulation MPC. In Fig. \ref{fig:po_comp}, we can see various solutions that the pose optimization block can compute to solve the same task based on the choice of weights for the objectives in the cost function. We can consider the pose that tracks COM height and pitch and only uses the arm to reach the object as the baseline. Then, if we reduce the weight related to the COM height in the pose optimization cost function, we obtain the pose in Fig. \ref{fig:po_squat}, where the controller trades the robot COM height to reach the object with the gripper. Similarly, in Fig. \ref{fig:po_pitch}, we can see the effect of reducing the weight related to robot body rotation. A balanced choice of weights in the cost function gives us a pose that uses all possible DOFs of the system to reach the object with the gripper, as illustrated in Fig. \ref{fig:po_mix}. All these starting poses can complete the task of lifting the object to a predefined height.

\subsection{Door Opening Using Proposed Approach}
\begin{figure}
    \medskip
    \centering
     \includegraphics[width=0.92\linewidth,trim=3pt 150pt 510pt 3pt, clip]{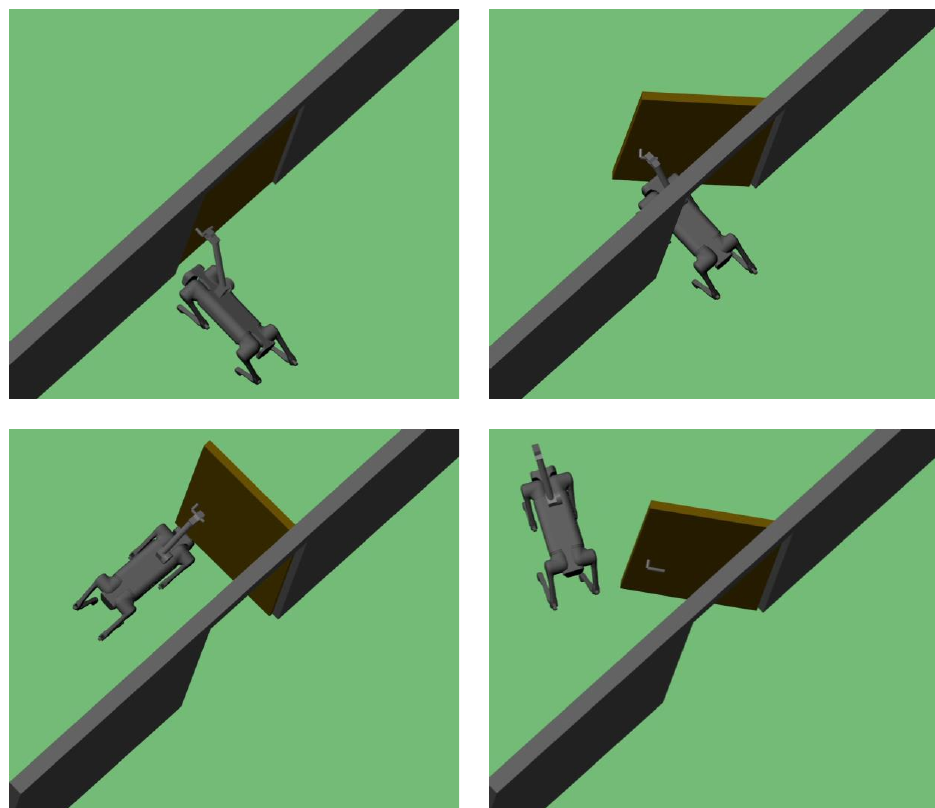}
    \caption{\textbf{Door opening with the proposed framework}. In the initial phase, the object manipulation planner and pose optimization dictate the robot to roll to manipulate the door handle. Subsequently, the robot applies the necessary force to push the door open, with the manipulation planner considering kinematic and spatial constraints.}
    \label{fig:door_opening}
\end{figure}

To illustrate the effectiveness of our approach, we conducted tests involving the challenging task of opening a heavy, resistant hinged door. Fig. \ref{fig:door_opening} presents snapshots of the task successfully solved in simulation. Within the manipulation planner, we have subtasks such as turning the door handle and pushing the door open. The pose optimization orchestrates the robot's motion to align with the desired manipulation actions. Notably, in this task, we can account for potential collisions with the door frame directly within the pose optimization, resulting in viable collision-free trajectories for the robot's COM.
One key advantage of our approach, with its distinct hierarchical components working together, is the ability to incorporate additional constraints into the problem without significantly increasing complexity and computational overhead. This is in contrast to a unified planner for both the object and robot, where the nonlinear structure would considerably complicate the process of obtaining an online planning solution.

\section{CONCLUSIONS}
\label{sec:conclusions}

This paper introduces a practical approach for addressing loco-manipulation challenges in legged robots. Our hierarchical approach simplifies the problem by breaking it into three components that work together effectively. We validate our approach with numerical and experimental scenarios investigating the effect of the three components separately. We show the importance of considering the object's dynamic effect on the robot controller, highlight the significance of our online object manipulation planner, and demonstrate the flexibility of our pose optimization component. Our framework, when applied to a Unitree Aliengo equipped with a custom-made robotic arm, achieves successful tasks such as lifting and carrying an 8 Kg object and opening a resistive hinged door, underscoring the significance of our three components. Future directions include extending the framework to more practical tasks to explore the boundaries of the current arm design.




\newpage
\bibliographystyle{IEEEtran}
\bibliography{IEEEexample}

\end{document}